\documentclass{article}

\usepackage{PRIMEarxiv}

\usepackage[utf8]{inputenc} 
\usepackage[T1]{fontenc}    
\usepackage{hyperref}       
\usepackage{url}            
\usepackage{booktabs}       
\usepackage{amsfonts}       
\usepackage{nicefrac}       
\usepackage{microtype}      
\usepackage{lipsum}
\usepackage{fancyhdr}       
\usepackage{graphicx}       
\graphicspath{{media/}}     

\usepackage{cite}
\usepackage{amsmath,amssymb,amsfonts}
\usepackage{algorithmic}
\usepackage{textcomp}
\usepackage{xcolor}
\usepackage{multirow}
\def\BibTeX{{\rm B\kern-.05em{\sc i\kern-.025em b}\kern-.08em
    T\kern-.1667em\lower.7ex\hbox{E}\kern-.125emX}}
\newcommand{\synp}{\textsc{SynQP}}

\title{SynQP: A Framework and Metrics for Evaluating the Quality and Privacy Risk of Synthetic Data 
}

\author{
  Bing Hu \\
  Computer Science \\
  University of Waterloo \\
  \texttt{b25hu@uwaterloo.ca} \\
   \And
  Yixin Li \\
  Public Health Sciences \\
  University of Waterloo \\
   \And
  Asma Bahamyirou \\
  Public Health Agency of Canada \\
  Government of Canada \\
   \And
  Helen Chen \\
  Public Health Sciences \\
  University of Waterloo \\
}

\begin{document}
\maketitle

\let\thefootnote\relax\footnotetext{22nd Annual International Conference on Privacy, Security, and Trust (PST2025), Fredericton, Canada}

\begin{abstract}
The use of synthetic data in health applications raises privacy concerns, yet the lack of open frameworks for privacy evaluations has slowed its adoption. 
A major challenge is the absence of accessible benchmark datasets for evaluating privacy risks, due to difficulties in acquiring sensitive data. 
To address this, we introduce \synp, an open framework for benchmarking privacy in synthetic data generation (SDG) using simulated sensitive data, ensuring that original data remains confidential. 
We also highlight the need for privacy metrics that fairly account for the probabilistic nature of machine learning models. 
As a demonstration, we use \synp to benchmark CTGAN and propose a new identity disclosure risk metric that offers a more accurate estimation of privacy risks compared to existing approaches. 
Our work provides a critical tool for improving the transparency and reliability of privacy evaluations, enabling safer use of synthetic data in health-related applications.
Our privacy assessments (Table II) reveal that DP consistently lowers both identity disclosure risk (SD-IDR) and membership-inference attack risk (SD-MIA), with all DP-augmented models staying below the 0.09 regulatory threshold.
Code available at \url{https://github.com/CAN-SYNH/SynQP}
\end{abstract}


\section{Introduction}

Despite great potential benefit through applied machine learning, access to sensitive personal information requires lengthy approval processes and often comes with stringent governing rules \cite{Kokosi2022AnOO, Dattani2013AccessingEA}.
Synthetic data is a privacy-enhancing technology (PET) that provides additional protection of sensitive information for data sharing and enables findable, accessible, interoperable, and reusable (FAIR) data standards \cite{Tsao2023HealthSD}. 
Generative artificial intelligence methods such as GAN \cite{Xu2019ModelingTD}, VAE \cite{Xu2019ModelingTD}, and DDPMs \cite{kotelnikov2023tabddpm} have shown great potential in generating high-quality synthetic data with high utility and fidelity while enhancing privacy protection when compared to the real data.

However, it is well known that machine learning models can overfit their training data. 
When overfitting occurs, a synthetic record may closely resemble a real individual from the original dataset.
As a result, if synthetic data is shared or released publicly, there is a risk that personal information could be inadvertently disclosed \cite{idreman}. 
Regardless of well defined and established metrics for evaluating the utility and fidelity of synthetic data \cite{hu2023evaluation}, there is a lack of established open framework and fair metrics for the evaluation of privacy risks, such as re-identification, membership attack and attribute inference attack \cite{idreman, hu2022membership}. 
Given the primary concern of synthetic data in health applications is privacy, the lack of an open framework for privacy evaluations hinders the adoption of synthetic data technologies \cite{Tsao2023HealthSD}.  
A key challenge for privacy evaluations is the lack of open data for benchmarking of privacy evaluation metrics for synthetic data generation (SDG) due to the difficulty to access original data \cite{el2013guide, idreman, gomes2022don}.
Open datasets are typically fully de-identified or contain minimal identifiable information, making them unsuitable to benchmark SDG methods for privacy risk evaluation \cite{johnson2023mimic}. 
\begin{figure}
    \centering
    \includegraphics[width=0.85\textwidth]{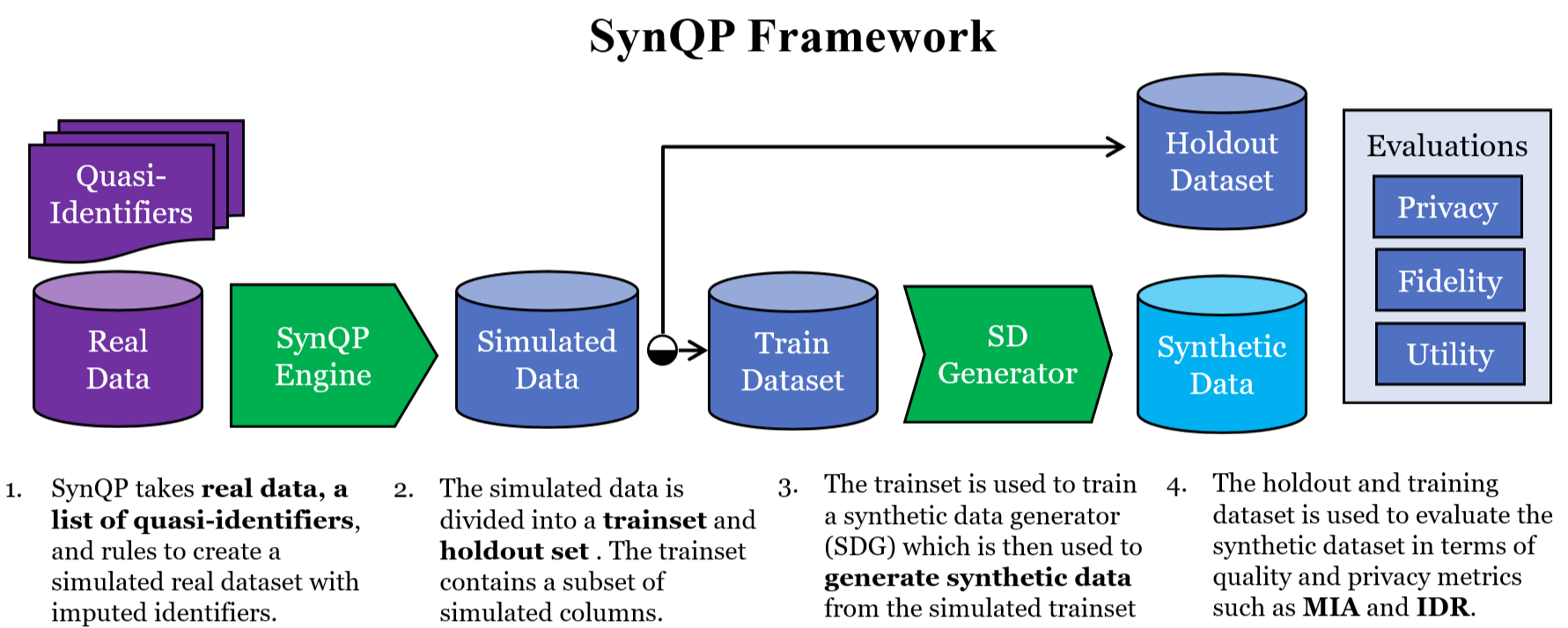}
    \caption{SynQP workflow and framework for simulated data generation and synthetic data evaluation. }
    \label{fig:spriv}
\end{figure}
Conversely, studies that conduct privacy evaluations for their SDG methods often rely on proprietary or private datasets containing identifying information which makes it challenging to compare and benchmark their SDG model against other models \cite{idreman}.
Consequently, SDG publications frequently omit privacy evaluations, despite privacy being a critical aspect, or a main motivation for synthetic data generation \cite{kotelnikov2023tabddpm, Solatorio2023REaLTabFormerGR, Xu2019ModelingTD}.  
Omitting such key indicators makes it hard for decision-makers for the adoption of synthetic data as a viable PET \cite{pushkarna2022data, gomes2022don}. 

A secondary challenge is the translational discordance between governance, privacy, and legal requirements to technical requirements and vice versa. 
Strict privacy regulation with vague legal language, complex institutional governance, and lengthy approval processes all contribute to highly risk-averse organizations and equally as challenging to define technical requirements \cite{goel2022expert, Tsao2023HealthSD}. 
A lack of common understanding and clear translation between complex policy requirements and technical requirements hinder AI innovation, slow research progress, and limits improvements in patient care \cite{Tsao2023HealthSD}. 

We introduce \synp, an open framework designed for benchmarking privacy evaluations of synthetic data using simulated pseudo-identifiable data constructed from non-identifiable real data.
This open framework enables AI researchers to benchmark their SDG models in privacy risk evaluation, making SDG models more comprehensive and actionable for policy- and decision-makers. 
Beyond benchmarking metrics, the \synp\ framework is also highly extensible and adaptable to different datasets, quasi-identifiers, and evaluation metics; enabling \synp\ to be used as an easy-to-use translational tool between policy and regulatory requirements to technical requirements and vice versa. 
By leveraging SDG as a PET, administrative data holders can safely liberalize their data, thereby accelerating research, collaboration and innovation. This is especially beneficial for AI innovations in healthcare.  

Our main contributions are as follows:

\begin{itemize}
    \item We propose a novel benchmark framework \synp\ standardizing the evaluation of privacy risk of synthetic data generation models.
    \item We define a novel concept of synthetic data identity disclosure risk and membership attack to more accurately evaluate the privacy risks of synthetic data.
    \item We demonstrate empirical evidence of applying \synp\ and synthetic data identity disclosure risk to CTGAN. 
\end{itemize}

\section{Methodology}
\label{sec:headings}

As shown in \autoref{fig:spriv}, the \synp\ framework generates a simulated population which a subset is used to train an SDG to generate synthetic data. 
The generated synthetic data along with the population can then be used for synthetic data privacy evaluations. 
Through simulated data, the primary aim of the \synp\ framework is to provide a privacy benchmark for SDG methods where previously none was available.

\subsection{Open \synp\ Framework}

The idea of \synp\ is to simulate identifiable data for real datasets that do not contain identifying data. 
By simulating quasi-identifiers for de-identified real datasets, \synp\ can generate simulated open datasets for which SDGs can be evaluated upon.

\synp\ initiates the simulation process by seeding a population with quasi-identifiers following real data distributions. 
Once the population is established with these simulated quasi-identifiers, non-identifying datasets for real use cases are then linked to each seed quasi-identifier row.
Through linking real use case non-identifiable data with seeded quasi-identifier data, we constructed our simulated pseudo-identifiable population and SDG training dataset.
Synthetic data from SDG models trained on the simulated data can then be evaluated using our proposed synthetic data privacy risk metrics and other synthetic data evaluation frameworks \cite{yuan2024multi}. 

\subsubsection{Quasi-Identifiers}

Quasi-identifiers are pieces of information that are not unique identifiers by themselves, but instead can be correlated together to create a unique identifier \cite{sweeney2002k, el2013guide}. 
For example, we collect and seed quasi-identifier distributions for age, gender, marital status, occupation, ethnicity, and address. 

As the first seed of our simulated population, we apply inverse transform sampling for age using distributions gathered from the census \cite{Government_of_Canada_2023}. 
Age values span between 0 to 99. As gender is correlated with age, we then conditionally sample gender given age for \textit{men+} or \textit{women+}.

For each sampled age and gender, we random sample for marital status, occupation, ethnicity, and address for each row. 
We apply random sampling on a collected list of 1154 occupations, 7 marital statuses, and 250 ethnicities for each row.
Random CA US addresses are generated using an available Python package \cite{Neosergio}, our framework is not limited to US addresses and other methods for random addresses can be easily applied. 
Generated random addresses include street and street numbers, city, state, and postal code. 
For our current work, we do not consider possible correlations between age and gender to occupations, marital statuses, addresses, and ethnicities.

\subsubsection{Linked Real Data}

We complete our simulated population data by linking non-identifiable real-use case data to our seeded quasi-identifier data. 
For demonstration we link diabetes data from the National Institute of Diabetes and Digestive and Kidney Diseases (NIDDK) \cite{turner2011niddk} and publicly available BMI data \cite{Ersever_2018}. 
The diabetes dataset is a non-identifiable dataset with data columns of the number of pregnancies, glucose concentration, diastolic blood pressure, skin thickness, insulin, and BMI, with age as the only quasi-identifier.
The BMI data is a non-identifiable dataset with data columns of gender, height, weight, and BMI. 

Taking our seed quasi-identifier population, we conditionally sample height, weight, and BMI given gender using the distributions defined in our BMI data for our simulated population. 
Using BMI and age in our simulated population, we find the nearest neighbour in the diabetes dataset to infill diabetes data columns for our simulated population.
We could also consider using other sampling methodology such as k-nearest neighbours (k-NN). 
By identifying the most similar records based on actual age and BMI values, k-NN preserves the multivariate dependencies of the original data without imposing strong parametric assumptions, ensuring that our simulated population closely mirrors the statistical distribution of the real dataset with minimal degradation \cite{peterson2009k}. 

\subsection{Implementing Differential Privacy}

Differential privacy (DP) is a method to provide anonymization guarantees for the data used in training ML models \cite{bonawitz, Ponomareva_2023}. 
There are a variety of DP methods to be used for a variety of ML models to protect against empirical privacy attacks. We utilize a local DP approach \cite{Ponomareva_2023}, similar to $d_\chi$ \cite{feyisetan2019privacyutilitypreservingtextualanalysis}, where noise is injected into the examples prior to training. Our DP noise mechanism is as follows

\begin{equation}
    DP(x,\epsilon) = (1-\epsilon)x + \epsilon \times Laplace( \mu_X, \sigma_X)
\end{equation}

Given $\epsilon \in [0,1]$ is the privacy budget, and $X\sim p_{data}$ with mean $\mu_X$ and standard deviation $\sigma_X$. 
For our experimentation and demonstration, we select $\epsilon$ levels of $0$ and $0.8$. 
Although $0.8$ is quite large for introduced noise, which is typically the case for input-level DP \cite{Ponomareva_2023}, this suffice for sake of demonstration and experimentation. 

\subsection{Quality Evaluations for Synthetic Data }

\subsubsection{Fidelity}
refers to the degree to which the synthetic data reproduce the statistical properties and underlying distributions of the original data. 
Hellinger distance (HD) is a common way to assess fidelity by comparing the probability distributions of individual features between synthetic and original data sets. Hellinger distance is expressed in eq:

\[ H^2(p, q) = \frac{1}{2} \sum_{i=1}^{n} \left(\sqrt{p_i} - \sqrt{q_i}\right)^2 \]
A result closer to 0 indicates that the two distributions are similar, while a result closer to 1 means that the two distributions are different. 
A common rule of thumb is that values below 0.1 are considered as excellent match, and  values below 0.2 are generally acceptable, while higher values indicate higher discrepancies.

\subsubsection{Utility}
refers to the usability of synthetic data in statistical modeling \cite{hu2023evaluation}. One such utility measurement is machine learning efficiency (MLE). 
The idea of MLE is to determine whether the synthetic data can be used as a replacement as the real data to develop statistical modeling \cite{basri2023hyperparameter}. 

In MLE, a proxy classification task is defined and two models are trained on the real data training set and the generated synthetic training set. 
Both models are then evaluated on a real-data test set where a highly capable SDG model should be on par with its real-data counterpart. 
In our context, we built a predictive logistic regression model trained on synthetic data and applied on simulated data. We measure performance results using AUC, where a higher AUC score indicates the higher utility of the synthetic data.

\subsection{Privacy Evaluations for Synthetic Data }

An open framework for evaluating the privacy of synthetic data necessitates the development of fair and robust privacy metrics that account for the probabilistic nature of machine learning models \cite{Tsao2023HealthSD}.  

De-identified data and synthetic data are generated using fundamentally different methodologies, each resulting dataset with distinctive characteristics. 
Whereas de-identification predominantly involves logical processes, synthetic data generation is primarily probabilistic. Due to this probabilistic nature of SDG, small variations in numerical data columns are likely to occur between very similar real and synthetic data. 

\begin{table*}[t]

\centering
\begin{tabular}{l|ccccccc|c}
\toprule
Model&Age&BMI&Pregnancies&\# Meds&\# labs&Time in Hosp.&\# Visits&\textbf{Avg.}\\
\midrule
CTGAN&\textbf{0.05}&0.11&0.27&\textbf{0.16}&\textbf{0.10}&\textbf{0.13}&\textbf{0.45}&\textbf{0.18}\\
CTGAN-DP&0.19&\textbf{0.12}&0.38&0.24&0.24&0.21&0.55&0.27\\
\midrule
TVAE&\textbf{0.40}&\textbf{0.20}&\textbf{0.39}&\textbf{0.20}&\textbf{0.51}&0.54&\textbf{0.16}&\textbf{0.34}\\
TVAE-DP&0.51&0.44&0.53&0.48&0.54&\textbf{0.47}&0.66&0.51\\
\midrule
GaussianCopula&\underline{\textbf{0.02}}&\underline{\textbf{0.05}}&\underline{\textbf{0.03}}&\underline{\textbf{0.14}}&\underline{\textbf{0.05}}&\underline{\textbf{0.11}}&\underline{\textbf{0.02}}&\underline{\textbf{0.06}}\\
GaussianCopula-DP&0.13&0.06&0.52&0.17&0.07&0.21&0.53&0.24\\
\bottomrule
\end{tabular}
\caption{Hellinger Distance and MLE Across Different Models. The best values with and without DP are bolded and the best values across all models are underlined.}
\label{tab:hddistances}
\end{table*}

\begin{figure*}[t]
\centering
\includegraphics[width=0.85\textwidth]{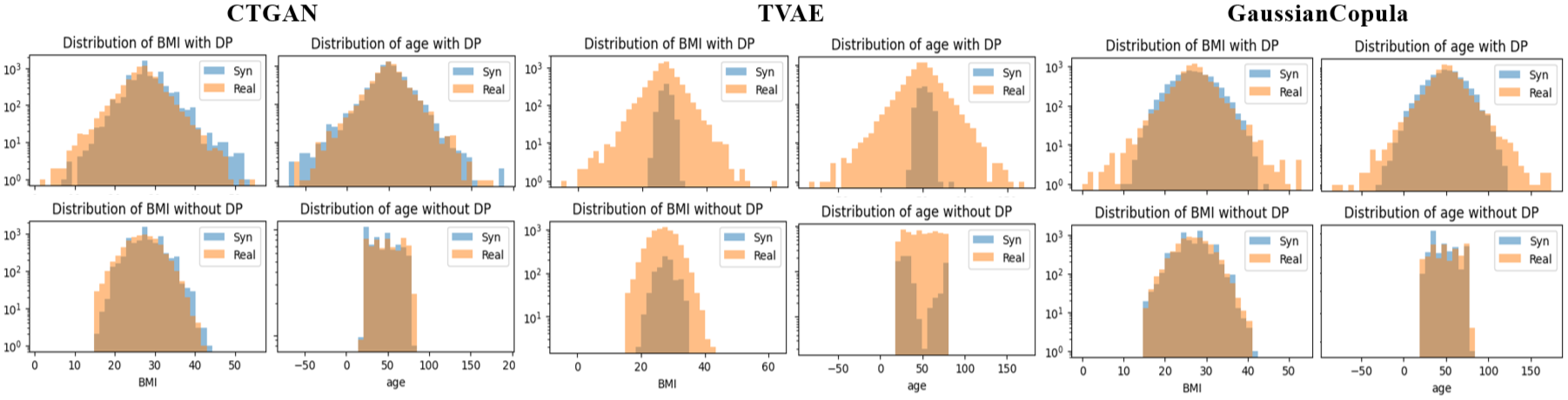}
\caption{Comparing training data corresponding generated synthetic data distributions for age and BMI with and without DP for CTGAN, TVAE, and GaussianCopula.}
\label{fig:univariate}
\end{figure*}

\subsubsection{Synthetic Data Identity Disclosure Risk (SD-IDR)}

Privacy metrics such as identity disclosure risk (IDR) \cite{idreman} that rely on cardinality of exact matches of numerical identifiers are not necessarily fair measures of the privacy risk of synthetic data generated using probabilistic models. 
Identity disclosure risk can be simplified to two parts: Real-to-Synthetic Identification Risk, and Synthetic-to-Real identification Risk \cite{idreman}.  
The IDR risk is expressed in eq.~\ref{idr} \cite{idreman}:

\begin{equation}
IDR = max\left( 
\frac{1}{N} \sum^n_{s=1}{\left(\frac{1}{f_s} * I_s\right)},
\frac{1}{n} \sum^n_{s=1}{\left(\frac{1}{F_s} * I_s\right)}
\right)
\label{idr}
\end{equation}

Where $N,n$ is the number of records in the real dataset and synthetic datasets respectively, $F_s, f_s$ is the size of the set of records with the same quasi-identifier values as record \textbf{s} in the real data and synthetic data respectively, and $I_s$ is the binary indicator of whether a record \textbf{s} in the real data \textit{exactly} matches a record in the synthetic data. 

Given the low likelihood of exact numerical matches between very similar real and synthetic data, IDR and other privacy metrics relying on exact matches potentially underestimate the privacy risk of synthetic data.
We propose a synthetic data identity disclosure risk (SD-IDR) that takes into account the variability of SDG models in producing small numerical variations. 
For SD-IDR, we propose a new definition for binary match indicator $I_s$ that takes into account small numerical variations. 
By accounting for these small numerical variations, we aim to resolve a research gap of creating a metric able to account for SDG stochasticity \cite{pilgram2025consensusprivacymetricsframework}. 
For SD-IDR, we define binary indicator $I_s$ as $1$ if a record \textbf{s} in real data and a record $SD_j$ in synthetic data matches exactly on categorical data columns and with cumulative differences less than $\epsilon$ on numerical data columns and $0$ otherwise. 




\subsubsection{Membership Inference Attack (MIA) \cite{shokri2017membershipinferenceattacksmachine}}
looks to determine whether a given sample comes from the training data for the particular synthetic data generator model. 
Let $X \sim p_{data}$ be a subset of data from a distribution of interest, then $X_{syn} \sim G(z; X, \theta)$ be a synthetic data distribution given a generator $G$ trained on $X$, noise $z$, and model $\theta$. 
In MIA, an attacker looks to identify a record $x \in X$ under a specified list of constraints and assumptions. 

Typical constraints \cite{hu2022membership, vanbreugel2023membershipinferenceattackssynthetic} assume knowledge of $X_{syn}$, some knowledge of $\hat{X} \sim p_{data}$ from subset of real data, and sometimes access to the generator $G$ and model parameters and gradients $\theta$. 
In cases where data publishers make available their models $G(X, \theta)$, past work \cite{shokri2017membershipinferenceattacksmachine, hu2022membership} has shown a variety of attacks, leveraging output probabilities and latent variables, to solve for training records. 
In most cases, these attacks can be avoided simply by not publishing the model. 

Without access to model parameters, instead we assume attackers only have access to $X_{syn}$. 
Being most conservative, let's assume the attacker has some $\hat{X} \sim p_{data}$ subset of the real data that they are interested in identifying in the real training data.
With this setup, MIA becomes a measurement of overfitting \cite{vanbreugel2023membershipinferenceattackssynthetic}; the attacker wishes to identify $x \sim X \cap \hat{X}$ of which the model $G(X,\theta)$ has overfitted. 

Thus, we define a straightforward synthetic data MIA (SD-MIA) as follows:
\begin{equation}
    MIA = IDR(X, X_{syn}) - IDR(X, \hat{X})
\end{equation}

Where $IDR(P,Q)$ is a measurement of the identity disclosure risk, including SD-IDR, between real data $P$ and sample distributions $Q$. 
Through \synp, our SD-MIA is adaptable to different adversarial sampling scenarios \cite{pilgram2025consensusprivacymetricsframework}; where $p_{data}$ can represent either a population-level distribution, or a distribution closer to the training data.
SD-MIA measures chance of identification of records through overfitting $IDR(X, X_{syn})$ minus the random chance of match with random sampling $IDR(X, \hat{X})$. 
When SD-MIA is close to $0$, this means that $X_{syn}$ is just as similar to $X$ as some random sample $\hat{X}$, such that the attacker cannot learn anything meaningful beyond random chance. 
When SD-MIA $>> 0$, means that $X_{syn}$ has reduced diversity with $X$ compared to random sample $\hat{X}$, such that attackers matching record to $X_{syn}$ has greater chance to be in $X$ compared to random chance. 
In the case when SD-MIA $<0$, the model is underfit and can undergo more training. A ratio $MIA = IDR(X, X_{syn}) : IDR(X, \hat{X})$ can be considered as well in the context of the baseline risk.

\section{Results: A Case Study}
Through \synp, we are able to simulate a pseudo-identifiable dataset from a de-identified diabetes dataset that can be used to evaluate the privacy risk of various SDG models. 
We demonstrate \synp\ by evaluating IDR and SD-IDR on synthetic data generated from CTGAN \cite{Xu2019ModelingTD}, TVAE, and GaussianCopula trained from our simulated population. 
Using \synp\ we simulated a population of 10000 rows for the diabetes dataset using our above methodology. 
From this simulated population, we sample 7000 rows for our training set with quasi-identifier columns age, gender, marital status, and data columns BMI, number of pregnancies. 
The rest of 3000 rows were kept as holdout datasets for further evaluation of privacy risk.
We train CTGAN, TVAE, and GaussianCopula models with default hyperparameters from the Python library Synthetic Data Vault \cite{patki2016synthetic} on our training set to generate 10000 synthetic data rows. 

\subsection{Quality Evaluations}

\autoref{tab:hddistances} compares the HD between real data and synthetic data across different generative models, CTGAN, TVAE, and GaussianCopula, both with and without DP.
The average HDs for CTGAN, GaussianCopula, and TVAE without DP are 0.18, 0.06, and 0.34 respectively. 
We see a tradeoff between fidelity and privacy where HD degrades when DP is added; with average HD values 0.27, 0.24, and 0.34 for CTGAN, GaussianCopula, and TVAE respectively. 
\autoref{fig:univariate} shows the distribution of training data and synthetic data with and without DP, we see that differential privacy greatly degrades the overall distribution of the generated synthetic data compared to the original training data. 

The MLE utility scores for CTGAN, GaussianCopula, and TVAE without DP are 0.97, 0.99, and 0.99 respectively.
There is similar tradeoff between utility and privacy where MLE degrades when DP is added; with MLE scores 0.47, 0.94, 0.87 for CTGAN, GaussianCopula, and TVAE respectively.

\begin{table*}[t]
\centering 
\begin{tabular}{lc|cccc}
\toprule
\multicolumn{1}{c}{}&\multicolumn{1}{c|}{} &\multicolumn{4}{c}{Privacy Metrics}\\
Model&IDR$\Delta$*&\multicolumn{2}{c}{{SD-IDR}} &\multicolumn{2}{c}{{SD-MIA}} \\
&&No DP& With DP&No DP&With DP\\
\midrule
\multirow{4}{*}{CTGAN} &0& 4.94e-02&3.45e-02& 2.76e-03&-3.21e-02\\
&1& 1.32e-01&8.94e-02& -2.11e-02&-7.05e-02\\
&2& 2.01e-01&1.34e-01&-3.21e-02&-1.08e-01\\
&3&2.47e-01&1.74e-01&-6.28e-02&-1.64e-01\\ 
\midrule
\multirow{4}{*}{TVAE} &0&6.44e-02&6.88e-02&8.88e-03&1.44e-02\\
&1&1.77e-01 &1.64e-01&2.77e-02&1.33e-02\\
&2&2.66e-01&2.40e-01&4.77e-02&1.00e-02\\
&3&3.44e-01&3.27e-01&5.22e-02&2.66e-02\\
\midrule
\multirow{4}{*}{GCop.} &0&4.55e-02&3.02e-02&3.34e-03&-9.71e-03\\
&1&1.31e-01&8.32e-02&-1.13e-02&-5.89e-02\\
&2&1.94e-01&1.30e-01&-1.97e-02&-8.27e-02\\
&3&2.45e-01&1.71e-01&-3.96e-02&-1.17e-01\\
\bottomrule
\end{tabular}
\caption{Privacy evaluations across models and differential privacy (DP) variants covering synthetic data identity disclosure risk (SD-IDR), and synthetic data membership inference attack (SD-MIA). *IDR$\Delta$ represents the variational budget set for SD-IDR, where $i$ represents the similarity threshold distance, with $i=0$ representing a threshold only for exact matches.
} 
\label{tab:privacyevals}
\end{table*}

\subsection{Synthetic Data Identity Disclosure Risk}

\autoref{tab:privacyevals} demonstrates the identity disclosure risk evaluations of different synthetic data generation models under different IDR variational budgets.
When computing the IDR for the generated synthetic data compared to our population data, the risk score for many combinations of variational budget and models are below the 0.09 threshold set by Health Canada and the European Medical Agency \cite{branson2020evaluating}.
Generally speaking, we observe that SD-IDR risks were lower with DP than compared to without DP. 
As the variational budget increases, we see corresponding increases in SD-IDR risk at different pace for different models, such that IDR$\Delta=0$ may underestimate the privacy risk. 
Therefore, multiple variational budgets including at 0 should be evaluated for a holistic representation of identity disclosure risk for synthetic data. 

\subsection{Synthetic Data Membership Inference Attack}


\autoref{tab:privacyevals} demonstrates the membership inference attack risk evaluations of different synthetic data generation models under different IDR variational budgets. 
We sample 3000 rows from our simulated population as $\hat{X}$, known to an attacker, to compute MIA. 
Generally speaking, we observe that SD-MIA risks were lower with DP than compared to without DP.
As the variational budget increases, we see a different trend for TVAE compared to CTGAN and GaussianCopula; where SD-MIA for TVAE increases as the variational budget increases, and CTGAN and GaussianCopula SD-MIA is decreasing to negative values as the variational budget increases.
This differing trend in terms of SD-MIA follows from \autoref{fig:univariate}, where we can see that TVAE has grossly overfit certain rows in the training data, whilst CTGAN and GaussianCopula may still be underfit and can undergo additional training given the negative SD-MIA values. 
Therefore, provided with multiple variational budgets, SD-MIA is a robust metric in detecting variance-bias tradeoff in terms of model underfitting or overfitting.



\section{Discussion}

In previous work, \synp\ has been applied to evaluate CTGAN as SDG model \cite{hu2024synprivacy}.
They have found that compared to SD-IDR, IDR could potentially underestimate re-identification risk. 
This study extends that work to prove that three common synthetic data generation models as well as differential private variants: CTGAN, TVAE, and GaussianCopular can be evaluated using our refined framework and quality and privacy evaluation metrics.

The rapid expansion in artificial intelligence generates vast volumes of real-world data (RWD) including electronic medical records and electronic health records. 
However, there are obstacles to accessing RWD due to privacy concerns about patient-level data leakage.
In this paper we show how the \synp\ framework allows the benchmarking and comparison on metrics evaluating both quality and privacy of different synthetic data generators with and without differential privacy. 
Our framework enables creation and sharing of open simulated datasets which can be used to by researchers to benchmark SDG methods in terms of both quality and privacy. 

Given the probabilistic nature of SDG methods, we define a synthetic data identity disclosure risk and a membership inference attack metric that better measures the risk of an attacker leveraging said probabilistic nature of synthetic data. 
We show how slightly modifying IDR to consider probabilistic attacks can greatly increase computed risk, showing how our SD-IDR produces an risk estimate that is more conservative. 
Additionally, we define a membership inference attack metric leveraging identity disclosure risk, which is able to provide guidance on whether a model is overfit or underfit. 
Through our experimentation with differential privacy, we show a tradeoff between utility, fidelity, and privacy. Compared to baseline, as DP increases, utility and fidelity both decrease alongside with privacy risk.  

Through combining results of quality and privacy metrics, an evaluation matrix can be constructed, where combinations of model and privacy budget that satisfy both quality and privacy regulatory requirements can be identified. 
The \synp\ framework can serve as a bridge between technical and policy and regulatory requirements by providing a standardized, transparent way to assess and quantify both data utility and privacy risk.
This quantification can be used to demonstrate compliance with regulatory thresholds and standards, while also informing policy decisions on acceptable trade-offs between data utility and privacy.
Given \synp's extensibility to additional quasi-identifiers and adaptability for any datasets, applying \synp\ makes it straightforward to define regulatory and contractual privacy requirements in terms of testable technical requirements and vice versa. 
Thus, \synp\ serves purpose of building common understanding between policy-makers, regulators, synthetic data builders and researchers, and downstream stakeholders. 



\section{Conclusion}

Through \synp, we are able to simulate a pseudo-identifiable dataset from any de-identified dataset that can be openly shared and used to evaluate the privacy risk of various SDG models. 
We demonstrate \synp using a diabetes dataset and with three most popular SDG models with and without differential privacy: CTGAN, TVAE and GaussianCopula. 
Additionally, we present a synthetic data identity disclosure risk and membership inference attack that better considers the stochastic nature of SDG models.
For future work, we will aim to further develop our \synp framework, apply our framework to additional models and datasets, and publish an extensive open dataset for future SDG benchmarking purposes. 

\bibliographystyle{unsrt}  
\bibliography{references}

\end{document}